# Enhancing Machine Learning for Imbalanced Medical Data: A Quantum-Inspired Approach to Synthetic Oversampling (QI-SMOTE)

Vikas Kashtriya

National Institute of Technology, Hamirpur, India, Vikas_phdcse@nith.ac.in

DoCSE, NIT Hamirpur, Himachal Pradesh, India 177005

Pardeep Singh

National Institute of Technology, Hamirpur, India, Pardeep@nith.ac.in

DoCSE, NIT Hamirpur, Himachal Pradesh, India 177005

## ABSTRACT

Class imbalance remains a critical challenge in machine learning (ML), particularly in the medical domain, where underrepresented minority classes lead to biased models and reduced predictive performance. This study introduces Quantum-Inspired SMOTE (QI-SMOTE), a novel data augmentation technique that enhances the performance of ML classifiers, including Random Forest (RF), Support Vector Machine (SVM), Logistic Regression (LR), k-Nearest Neighbors (KNN), Gradient Boosting (GB), and Neural Networks, by leveraging quantum principles such as quantum evolution and layered entanglement. Unlike conventional oversampling methods, QI-SMOTE generates synthetic instances that preserve complex data structures, improving model generalization and classification accuracy. We validate QI-SMOTE on the MIMIC-III and MIMIC-IV datasets, using mortality detection as a benchmark task due to their clinical significance and inherent class imbalance. We compare our method against traditional oversampling techniques, including Borderline-SMOTE, ADASYN, SMOTE-ENN, SMOTE-TOMEK, and SVM-SMOTE, using key performance metrics such as Accuracy, F1-score, G-Mean, and AUC-ROC. The results demonstrate that QI-SMOTE significantly improves the effectiveness of ensemble methods (RF, GB, ADA), kernel-based models (SVM), and deep learning approaches by producing more informative and balanced training data. By integrating quantum-inspired transformations into the ML pipeline, QI-SMOTE not only mitigates class imbalance but also enhances the robustness and reliability of predictive models in medical diagnostics and decision-making. This study highlights the potential of quantum-inspired resampling techniques in advancing state-of-the-art ML methodologies.

## 1 INTRODUCTION

### 1.1 Background

Class imbalance presents a significant challenge in many real-world ML scenarios, where one class of data is substantially larger than the other(s). This uneven distribution can cause models to be biased towards the majority class, often neglecting the minority class. This bias can degrade classification performance, particularly for the minority class, which is usually of greater interest in various applications, including medical contexts [1]. Medical datasets, often characterized by a wide range of features, including clinical variables and social determinants of health (SDOH), can further complicate this challenge.

When dealing with medical data, the issue of class imbalance is very common. For instance, in case of adverse drug events (ADEs) research, the instances of adverse events are typically much lower than the other biomedical events. When ML models are trained on such imbalanced datasets, they might achieve high accuracy by merely predicting the majority class, thereby overlooking the critical minority instances, which in this context, could be potential disease cases [2]. When trained on such skewed datasets, ML models may show high overall accuracy by predominantly predicting the majority class. This approach can lead to the critical oversight of minority instances, which in medical scenarios could represent actual cases of the disease [2]. The consequences of incorrectly classifying these minority instances are often grave, particularly in healthcare where early detection and treatment are vital. The challenge is not only to achieve high overall accuracy but to ensure the model is adequately sensitive to the subtleties of the minority class [3].

Various strategies have been proposed to counteract class imbalance, including data-level methods like oversampling and undersampling, algorithmic adjustments, and cost-sensitive learning [4]. Among these, the Synthetic Minority Over-sampling Technique (SMOTE) has become prominent for its effectiveness in generating synthetic samples for the minority class, aiding in balancing class distribution [5]. However, despite the success of SMOTE and similar methods, there is still a pressing need for more sophisticated techniques. These advanced methods should be capable of effectively handling the complexities of high-dimensional, multi-label medical data, ensuring that models are not only accurate but also equitable in their classification, particularly in critical fields like healthcare where the stakes are high.

## 1.2 Motivation

The exponential increase in data across various fields has led to a critical challenge in ML, where efficiently analyzing and extracting valuable insights is often hindered by imbalanced class distributions. Despite numerous existing strategies designed to mitigate this problem, each technique has inherent limitations that restrict its effectiveness across different ML applications. Traditional resampling methods, such as random oversampling and undersampling, attempt to equalize class distribution by duplicating minority instances or removing majority instances. However, these approaches can introduce significant drawbacks in ML models, such as overfitting in the case of oversampling or loss of critical information in undersampling, which ultimately weakens the classifier's performance [6]. More advanced techniques like SMOTE and its variations have gained popularity for generating synthetic minority class samples, which aid in improving ML model training. However, SMOTE is not without flaws, as it can lead to over-generalization, where synthetic samples fail to capture the true nature of the data distribution [7]. One example of this flaw is the generation of unrealistic data, such as a pair of values showing a systolic blood pressure of 160 mmHg and a heart rate of 40 bpm. In reality, when a patient's heart rate is as slow as 40 bpm, the systolic blood pressure cannot be as high as 160 mmHg. This issue becomes more pronounced in high-dimensional datasets, such as those used in medical applications, where the synthetic data may blur class boundaries, thereby reducing the discriminative power of classifiers [8].

Beyond data resampling, algorithmic modifications aimed at increasing the sensitivity of ML models to the minority class often require complex hyperparameter tuning and may lack universal applicability across different classifiers [9]. Cost-sensitive learning, another widely explored approach, assigns varying misclassification costs to different classes. While effective in certain domains, determining an optimal cost matrix remains a domain-specific challenge that limits its adaptability to diverse ML problems [10]. Given these challenges, there is an urgent need for innovative ML solutions that not only address class imbalance but also enhance model robustness, generalizability, and predictive accuracy.

## 1.3 Objective

This study focuses on evaluating the effectiveness of QI-SMOTE in enhancing ML classification performance for imbalanced datasets, particularly in domain-specific applications, such as medical diagnostics. Instead of developing a new classification algorithm, this research aims to improve existing ML models by generating more representative synthetic data, reducing class imbalance, and ultimately improving classification performance. Additionally, the study provides a comparative analysis of QI-SMOTE against recently introduced oversampling techniques to assess its impact on real-world datasets. The key contributions of this research can be summarized as follows:

- Introducing QI-SMOTE, a quantum-inspired resampling technique, to effectively manage class imbalance in ML tasks.
- Evaluating the impact of QI-SMOTE on various ML classifiers, including Random Forest, Support Vector Machine, Logistic Regression, Gradient Boosting, and k-Nearest Neighbors, across datasets with severe class imbalance.

- Comparing QI-SMOTE with traditional oversampling methods, such as SMOTE, Borderline-SMOTE, ADASYN, and SMOTE-TOMEK, to assess its effectiveness in improving classification accuracy, model robustness, and generalization.

## 2 RELATED WORK

SMOTE has been a revolutionary approach to addressing the problem of class imbalance [3]. This method is centered around the creation of synthetic samples for the underrepresented class by interpolating between existing examples. Specifically, SMOTE generates new instances by identifying k-nearest neighbors for each instance in the minority class and then producing new samples along the lines that connect these instances to their neighbors. It accomplishes this by modifying the feature vector of a minority instance, using a weighted difference from one of its neighboring instances. While SMOTE was innovative, it faced limitations, particularly in over-generalization and increased class overlap, leading to poorer model generalization [14]. To overcome these limitations, various SMOTE extensions have been developed:

- Borderline-SMOTE (B-SMOTE) targets instances near the decision boundary. Although effective in identifying challenging samples, it might overlook valuable insights from stable minority instances [15].
- ADASYN (Adaptive Synthetic Sampling) varies the number of synthetic samples based on each minority instance's classification difficulty. However, it risks overfitting by generating more samples in already challenging areas [16].
- SMOTE-ENN combines SMOTE with the Edited Nearest Neighbor (ENN) rule for removing misclassified instances. However, this could also lead to losing important data near class borders [17].
- SMOTE-TOMEK employs Tomek links with SMOTE for clarifying class boundaries by removing overlapping samples. This approach might inadvertently delete crucial minority instances [18].
- SVM-SMOTE uses Support Vector Machines to generate synthetic samples in critical support regions. While effective, it can be computationally demanding and less efficient in high-dimensional spaces [19].
- MOSS selects candidates for new example generation considering their sensitivity to class imbalance, addresses the random candidate selection issue in traditional SMOTE [20].
- Radius-SMOTE is aimed at reducing class overlap, but it can sometimes generate synthetic samples that do not adequately represent the original data distribution [21].
- DeepSMOTE, combines deep learning with SMOTE, however, it might generate images that overly simplify complex class distributions [22].
- Cluster-based oversampling method designed for hepatocellular carcinoma patients offers a tailored approach but could potentially lead to overfitting due to its synthetic sample generation and may also present challenges in implementation due to its complexity [23].
- RFMSE algorithm integrates M-SMOTE and ENN with Random Forest for medical data, while enhancing patient identification, it may incur higher computational demands and requires careful parameter tuning to ensure optimal performance [24].
- SMOTE-PSOEV integrates particle swarm optimization and Egyptian vulture methods for optimized synthetic sample generation, however, it can be computationally intensive and may overfit on certain datasets [25].
- IF-RVFL combines SMOTE with semi-supervised learning but may face difficulties in accurately capturing complex disease patterns in liver disease detection [26].
- The hybrid sampling approach, combining LR-SMOTE and CCR, can be prone to oversampling and may not effectively address the inherent complexity of datasets [27].
- AWSMOTE applies SVM-based adaptive weights, but it might not adequately address the issue of minority class representation in high-dimensional spaces [28].

In addition to SMOTE and its various extensions, there are diverse array of methods to tackle the class imbalance challenge. These techniques can be broadly classified into several categories, each offering unique approaches to address this issue: Resampling Techniques (oversampling/undersampling) [29], Algorithm-level Approaches [30], Ensemble Methods [31], Cost-sensitive Learning [32] and Anomaly Detection Techniques [33]. While each of these techniques has its advantages, they also come with their set of challenges. For instance, simple resampling can lead to overfitting (in the case of oversampling) or loss of information (in the case of undersampling). Algorithm-level approaches require modifying existing algorithms, which might not always be feasible. Ensemble methods, while effective, can be computationally intensive. Cost-sensitive learning requires careful tuning of misclassification costs, and anomaly detection techniques might not always be applicable, especially when the minority class doesn't represent anomalies. Apart from traditional SMOTE variants, several oversampling approaches have been proposed to handle complex, nonlinear data structures more effectively. Kernel-SMOTE applies kernel transformations to map data into higher-dimensional spaces where class boundaries become more separable before applying SMOTE interpolation. This helps preserve nonlinear separability but introduces sensitivity to kernel choice and parameter tuning. Similarly, variational autoencoder-based oversampling (e.g., VAE-SMOTE) learns latent representations of minority class data and generates synthetic points within this learned manifold, capturing complex distributions with reduced overlap. Other approaches like LLE-SMOTE or t-SNE-assisted resampling attempt to approximate intrinsic data manifolds prior to interpolation. While these methods enhance geometric fidelity, they may suffer from interpretability issues or high computational costs, particularly on large, high-dimensional datasets like MIMIC-III [40-42] and MIMIC-IV [42-44].

Recent research in quantum machine learning has also explored quantum-augmented data preprocessing and representation. For example, studies have proposed using data-driven Hamiltonians or quantum kernels to inform classification boundaries and learn structural information from quantum state encodings. Quantum-enhanced generative models have also been applied to synthesize data distributions that preserve entanglement-based correlations. Entanglement models the idea that some features in medical data are tightly linked, for example, heart rate and blood pressure often change together. By simulating entanglement, we ensure that when one feature changes during sample generation, its correlated partner reflects a consistent and realistic adjustment, preserving natural dependencies. However, these methods have primarily focused on classification or feature mapping, rather than synthetic sample generation. The few recent attempts at quantum-inspired oversampling, such as Mohanty et al. [36], do not fully exploit layered entanglement or quantum evolution. In contrast, our QI-SMOTE framework introduces a novel layered entanglement architecture and utilizes the Variational Quantum Eigensolver (VQE) to optimize feature interaction before integrating with classical SMOTE. This makes QI-SMOTE a hybrid approach that combines the strengths of quantum state modeling and classical interpolation for more robust synthetic data generation.

## 3 QUANTUM-INSPIRED SMOTE (QI-SMOTE)

### 3.1 Conceptual Framework

Our method enhances the quality of synthetic samples for imbalanced datasets by incorporating principles inspired by quantum computing. First, we map each data point into a high-dimensional feature space using quantum-inspired transformations such as superposition and entanglement. These transformations help preserve complex relationships between features that might otherwise be lost during oversampling. Next, we use an optimization routine, adapted from the VQE, to refine these transformed representations. Finally, we perform classical interpolation (as in SMOTE), but within this enriched feature space, to generate realistic minority class samples that maintain the underlying data manifold. This hybrid process improves the structure and diversity of synthetic data while reducing the likelihood of generating artifacts.

Quantum computing represents a substantial shift from conventional classical computing, utilizing quantum mechanics principles to achieve processing feats that classical computers can't match [34]. At the heart of this technology are quantum bits, or qubits, which are distinct from traditional binary bits. Unlike classical bits that are strictly binary (0 or 1), qubits operate in a state of superposition, enabling them to represent both 0 and 1 at the same time. Superposition in our framework refers to encoding each data point in such a way that it simultaneously represents multiple potential feature configurations. We can think of it as placing the data in a probabilistic blend of states, like how a medical symptom might point to multiple underlying causes, not just one. This enables a richer representation of uncertainty and feature interactions. This capability of qubits is a significant aspect which makes quantum approaches very unique. It leads to effective management and analysis of complex, high-dimensional data, potentially revolutionizing how we approach challenges in data processing and ML [35]. While there are two recent works attempting to deploy a quantum approach to SMOTE [36, 47], they have not used quantum entanglement, which is a core concept of quantum computing. Building on the foundations of quantum computing, QI-SMOTE leverages key quantum principles [37], particularly superposition, entanglement and evolution, to address challenges in classical data processing techniques:

- Superposition in QI-SMOTE: Leveraging the principle of superposition [38], QI-SMOTE transforms the

typical linear combinations of features used in standard data processing. By allowing the representation of multiple feature states simultaneously, QI-SMOTE can process a broader range of data possibilities, enhancing the diversity and representativeness of the synthetic samples.

- Layered Entanglement in QI-SMOTE: Entanglement [39] ensures that the relationships within data are preserved, even when generating new samples. We implement a layered entanglement approach using a series of quantum gates. Starting with the CNOT gate to establish initial pairwise connections, we then intensify these links with Controlled-Z (CZ) and Toffoli gates, ensuring that synthetic samples are not just diverse but also accurately representative of the minority class.
- Evolution in QI-SMOTE: Quantum Evolution plays a crucial role in dynamically optimizing the quantum states towards the most effective synthesis of new data points. By employing methods like the Variational Quantum Eigensolver [39], QI-SMOTE evolves the entangled states in a controlled manner to minimize the energy configurations, ensuring that the synthetic samples generated are not only diverse and representative but also optimized for specific analytical objectives.

By integrating these quantum principles with the foundational concepts of SMOTE, QI-SMOTE aims to create a more robust and versatile oversampling technique. The quantum-inspired approach not only enhances the diversity of the generated synthetic samples but also ensures that they are representative of the actual data distribution. This could lead to substantial improvements in ML models trained on these balanced datasets, as it addresses key limitations of conventional oversampling methods, including the risk of overfitting and loss of crucial information. Consequently, QI-SMOTE stands out as a promising solution for enhancing data processing, particularly in fields where class imbalance is a significant issue. The pictorial representation of our conceptual framework is shown in Figure 1.

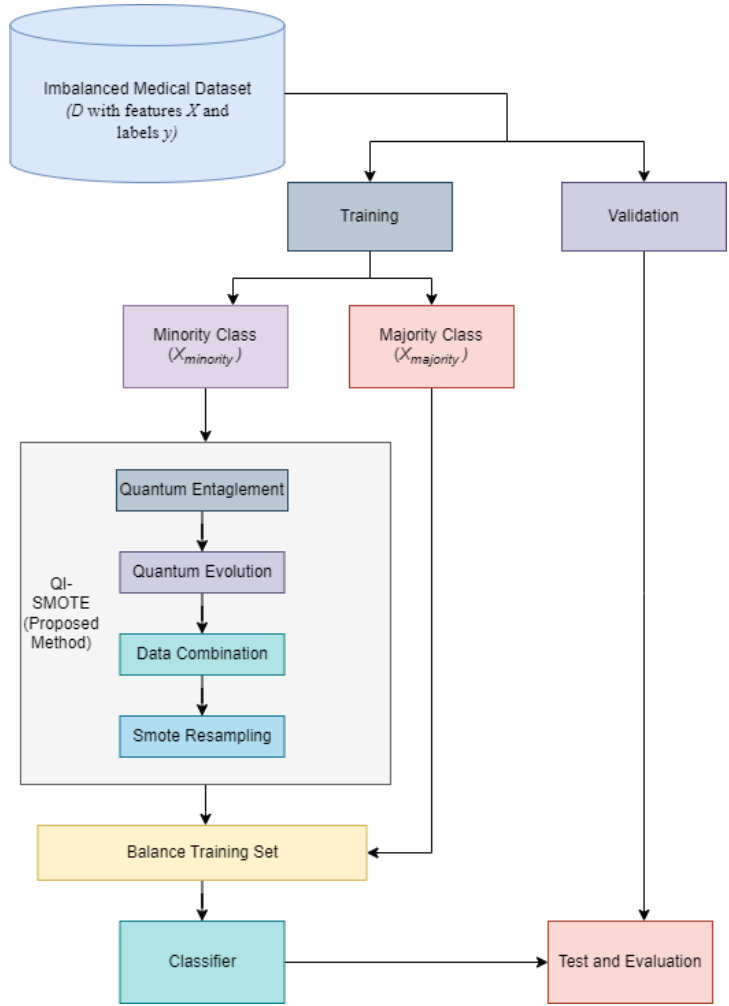

**Figure 1.** Conceptual Framework of QI-SMOTE Approach

## 3.2 Algorithm Details

The QI-SMOTE algorithm is designed to generate synthetic samples by leveraging key quantum principles specifically superposition, entanglement, and evolution alongside the foundational concepts of SMOTE. The process begins with the initialization of the minority class dataset, where the number of synthetic samples to be generated and specific quantum parameters, such as superposition coefficients and entanglement factors, are defined. The core of the algorithm involves two quantum-inspired processes: quantum superposition-inspired sample generation and quantum entanglement-inspired sample correlation. In the former, each minority class sample is processed to compute a superposition state based on its feature values and those of its k-nearest neighbours. This leads to the generation of a synthetic sample influenced by this superposition state. Meanwhile, the entanglement process involves computing an entanglement factor for pairs of synthetic samples, ensuring their feature values maintain a coherent relationship and preserve the underlying data distribution. In the QI-SMOTE algorithm, feature values from the minority class are encoded into quantum states using rotational gates. Specifically, each feature $f_i$ of a sample is mapped to a rotation angle in the RY gate, written as $RY(f_i)$. Intuitively, this process can be understood as treating each feature like a volume knob: the higher its value, the more it rotates a virtual qubit away from its initial position. This allows the quantum circuit to encode continuous variations in the data into the angular configuration of quantum states, increasing the expressiveness of the synthetic instance space. After rotation, entanglement gates (CNOT, CZ, and Toffoli) are applied in layers to connect these qubits. These gates model interdependencies between features, for instance, how heart rate and blood pressure might co-vary in a patient. This "layered entanglement" ensures that synthetic samples do not treat features in isolation but instead preserve the relational structure of real data. The degree of interaction is governed by entanglement factors, which act like "weights" determining how tightly two features are linked when generating a new sample. These interactions are fine-tuned using quantum evolution (via VQE) to ensure that the final synthetic points are not only diverse, but also representative of the underlying data distribution.

**Algorithm: QI-SMOTE Algorithm**

**Input:**
- Dataset $D$ with features $X$ and labels $y$.
- Minority class samples $X_{minority}$.

**Output:**
- Resamples dataset with features $X_{res}$ and labels $y_{res}$.

1. **Layered Quantum Entanglement**
   1.1. **For each sample $s$ in $X_{minority}$:**
      1.1.1. Initialize a quantum circuit with $n$ qubits, where $n$ is the number of features in $s$.
      1.1.2. Apply a Hadamard[†] gate to each qubit.
      1.1.3. For each feature $f_i$ in $s$, apply a rotation[†] $RY(f_i)$ to the corresponding qubit.
      1.1.4. Apply a CNOT[†] gate to entangle the qubits.
      1.1.5. Apply a CZ[†] gate to entangle the qubits.
      1.1.6. Apply a Toffoli[†] gate to entangle the qubits.
      1.1.7. Simulate the quantum circuit to obtain the entangled state vector $|\Psi\rangle$.
2. **Quantum Evolution:**
   2.1. **For each entangled state $|\Psi\rangle$:**
      2.1.1. Construct the Hamiltonian matrix $H$ using the outer product:
      $$H = |\Psi\rangle\langle\Psi|$$
      2.1.2. Use the VQE[†] to find the evolved state $|\Psi_{evolved}\rangle$. VQE will minimize the expectation value of $H$ with respect to a parameterized quantum state.
      2.1.3. Extract the real part of the evolved state to obtain $|\Psi_{real}\rangle$
3. **Data Combination**
   3.1. Append $|\Psi_{real}\rangle$ to the original dataset $D$.
   (Quantum-enhanced encoding complete, now applying classical SMOTE on enriched feature space)
4. **SMOTE Resampling:**
   4.1. Apply the traditional SMOTE algorithm to the combined dataset to generate synthetic samples.
5. **Return** the resampled dataset $X_{res}, y_{res}$.

[†] Detailed mathematical definitions and formulations of these quantum operations are provided in Appendix A.

On the completion of QI-SMOTE Algorithm step number 3, the quantum-inspired feature augmentation is complete. The original minority samples have been transformed through superposition, entanglement, and evolution into enriched synthetic vectors that encode complex feature interdependencies. With these quantum-enhanced representations now available, we transition back to the classical domain. The QI-SMOTE pipeline resumes with traditional SMOTE, which interpolates between both original and evolved synthetic samples to generate the final balanced dataset. This hybrid design allows us to leverage quantum principles to enrich feature diversity while maintaining compatibility with well-established oversampling frameworks. By incorporating quantum principles, QI-SMOTE aims to produce synthetic samples that are not only diverse but also representative of the actual data distribution, enhancing the robustness of ML models trained on such balanced datasets. This unique approach signifies a step forward in tackling the challenges of class imbalance, particularly in complex data environments like medical datasets.

## 4 EXPERIMENTAL SETUP

### 4.1 Datasets

In assessing the effectiveness of the QI-SMOTE algorithm, we have selected two highly regarded medical datasets, MIMIC-III and MIMIC-IV, known for their complexity, multidimensionality, and multi-label attributes:

- MIMIC-III (Medical Information Mart for Intensive Care III) [40-42]: This extensive and publicly accessible database comprises anonymized health records from over 40,000 patients admitted to the critical care units of Beth Israel Deaconess Medical Center from 2001 to 2012. It encompasses a diverse array of data types, including patient demographics, vital signs, lab test results, procedures, medications, caregiver notes, imaging reports, and mortality data. MIMIC-III's complexity mirrors the multifaceted nature of real-world healthcare scenarios, making it a valuable resource for testing and refining computational approaches that aim to improve patient outcomes and healthcare decision-making.
- MIMIC-IV (Medical Information Mart for Intensive Care IV) [42-44]: As an advancement of MIMIC-III, MIMIC-IV offers a more comprehensive dataset by including more recent patient admissions and a broader set of features. It continues to provide detailed information on patients in critical care settings, thereby serving as a valuable resource for developing and validating ML models specifically designed for healthcare applications.

The selection of these datasets is driven by their inherent complexities and class imbalances, which are representative of many real-world medical data challenges.

### 4.2 Experimental Design

The preparation of MIMIC datasets for the analysis of resampling techniques involved a comprehensive process, starting from the selection of a classification task to the final evaluation of the techniques. For the selection of classification task, the mortality detection was chosen because of its widespread popularity in recent benchmark studies involving MIMIC datasets [45].

To begin, data was extracted and preprocessed from various CSV files present in the MIMIC datasets. Key features were selected based on their relevance and impact on the mortality detection task [46]. These features included "gender", "anchor_age", "admission_type", "admission_location", "insurance", "heart_rate", "systolic_bp", "diastolic_bp", and "icd_codes." The choice of these features was guided by their established significance in recent benchmarks of the MIMIC dataset. The MIMIC-III and MIMIC-IV datasets were carefully curated, focusing on specific features. A key characteristic of these datasets is their class imbalance, measured using the Imbalance Ratio (IR), which compares the count of majority class instances to minority class instances. The IR for MIMIC-III is 2.78, indicating a notable imbalance, and for MIMIC-IV, it is slightly lower at 2.05, showing different levels of imbalance in these datasets.

To investigate how class imbalance affects various resampling methods, modified versions of both datasets were created. These versions, named MIMIC-III (10), MIMIC-III (20), MIMIC-IV (10), and MIMIC-IV (20), were designed with higher IRs of 10 and 20 by selectively removing samples from the minority class. This was crucial for testing how well resampling methods perform under more pronounced class imbalances. For evaluating these methods, the datasets were divided into training and test sets. The training set, reflecting the original imbalances, was altered using resampling strategies that either increased the minority class, decreased the majority class, or both. Various classifiers were trained for mortality prediction on these balanced training sets. The effectiveness of the resampling techniques was then assessed based on how these classifiers performed on the unaltered test set. This allowed for an evaluation of the models' ability to generalize to new, unbalanced data. The process involved comparing performance metrics like accuracy, precision, recall, F1-score, and AUC-ROC across models and techniques. This comparison provided insights into how each resampling technique manages class imbalance, especially for mortality detection in the MIMIC datasets. This analysis is vital for understanding the challenges of class imbalance in medical datasets and the effectiveness of resampling techniques in addressing these issues. This emphasizes the importance of tailored data pre-processing in improving the performance of ML models in healthcare, where accurate predictions can have significant implications for patient outcomes.

### 4.3 Baseline Methods

For a comprehensive assessment of QI-SMOTE, it's essential to benchmark it against established data balancing techniques. This comparative analysis will shed light on QI-SMOTE's relative strengths and areas for improvement. We will compare QI-SMOTE with the following baseline methods:

1. SMOTE [3]: As the basis for QI-SMOTE, comparing it with the original SMOTE algorithm is crucial. SMOTE generates synthetic samples in feature space to equalize class distribution.
2. ADASYN (Adaptive Synthetic Sampling) [16]: This adaptive version of SMOTE produces synthetic data based on density disparities between classes, focusing on harder-to-classify samples.
3. Borderline-SMOTE [15]: This SMOTE variant concentrates on minority class borderline instances to refine the decision boundary between classes.
4. Random Over-sampling [32]: A simple method that duplicates minority class instances to balance classes.
5. Random Under-sampling [32]: This approach randomly removes majority class instances but risks losing valuable information.
6. SMOTE-ENN [17]: Combines SMOTE with the Edited Nearest Neighbor rule for oversampling and data cleaning.
7. SMOTE-TOMEK [18]: Integrates SMOTE with Tomek links to eliminate overlaps post-oversampling.
8. SVM-SMOTE [19]: Employs SVM to identify support vectors for the minority class for synthetic sample generation.

Comparing QI-SMOTE against these methodologies will highlight its advantages and showcase where it excels over traditional techniques, especially in the complex arena of multidimensional medical data analysis.

### 4.4 Evaluation Metrics

To conduct a comprehensive assessment of the QI-SMOTE, especially in handling imbalanced medical datasets, we will apply a range of established evaluation metrics. These metrics are crucial for not only assessing the algorithm's accuracy in classifying instances but also for evaluating its proficiency in addressing the challenges of the minority class, which is vital in imbalanced datasets:

1. F1-score: This metric, which represents the harmonic mean of precision and recall, provides a balanced measure of the algorithm's ability to minimize false positives and false negatives. A higher F1-score, approaching 1, indicates more effective classification, which is particularly important in medical contexts where inaccuracies can have serious implications.
2. G-Mean: Calculated as the geometric mean of sensitivity (true positive rate) and specificity (true negative rate), the G-Mean is especially useful in imbalanced dataset scenarios. It evaluates the algorithm's performance evenly across both classes, with a higher G-Mean suggesting a more balanced and effective classification across minority and majority classes.
3. AUC-ROC: This metric evaluates the capability of a binary classifier to differentiate between classes. An AUC (Area Under the Curve) value close to 1 denotes an excellent classifier, whereas a value around 0.5 indicates a lack of discriminatory ability.
4. Precision, Recall, and Accuracy: While the F1-score and G-Mean offer integrated insights, analyzing precision (the proportion of true positives among positive predictions), recall (the proportion of true positives identified among actual positives), and accuracy (the proportion of true results, both true positives and true negatives, in the dataset) separately provides a deeper understanding of the model's specific strengths and weaknesses.

## 5 RESULTS AND DISCUSSIONS

In our experiments, we employed a diverse set of resampling techniques to address the class imbalance problem, with a particular focus on the QI-SMOTE method. The following sections detail our observations and insights derived from an in-depth analysis of the datasets and classifier performance.

### 5.1 Data Distribution after Sampling

In our initial exploration, we applied QI-SMOTE and various baseline algorithms to a synthetically generated binary dataset. This dataset was created using random numbers drawn from normal distributions and a standard normal distribution, specifically to study the impact of these techniques on data distributions. The resulting bidimensional scatter plots, as illustrated in Figure 2, reveal the unique characteristics imparted by each sampling method on the dataset.

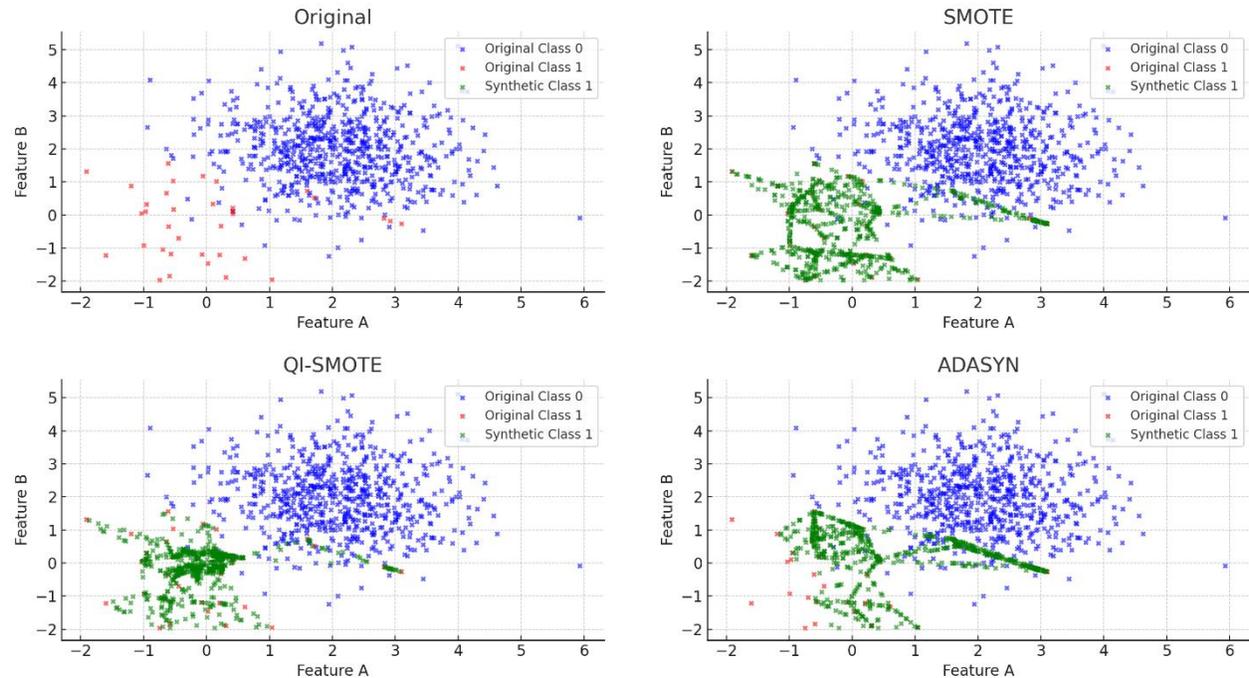

**Figure 2.** Visualization of synthetic sample distributions in a 2D feature space generated by various resampling techniques, including QI-SMOTE. Each plot highlights the spatial distribution of minority and synthetic samples relative to the majority class.

The original dataset, characterized by a clear demarcation between the classes, vividly highlighted the prevalent issue of class imbalance. This visualization underscored the necessity of employing resampling techniques to address this imbalance. Within this framework, the QI-SMOTE approach demonstrated its effectiveness. It displayed synthetic samples that were strategically interspersed among the original data points, effectively bridging the gap between the classes. This was achieved without causing an overshadowing of the original data, thereby maintaining the dataset's integrity.

Conversely, methods such as Random Oversampling and Undersampling displayed their typical behaviors. Random Oversampling works by duplicating instances in the minority class, thereby increasing its representation. In contrast, Undersampling reduces the number of instances in the majority class. Meanwhile, SMOTE and its derivatives create synthetic samples in areas where the minority and majority classes converge. Nonetheless, the patterns in which these samples are distributed differ slightly but significantly from those produced by QI-SMOTE, especially in terms of the complexity of the resulting dataset These differences in sample distribution are crucial. Random Oversampling, while simple, can lead to overfitting due to the repeated minority class instances. Undersampling, though effective in balancing classes, risks losing valuable information by removing instances from the majority class. SMOTE and its variants attempt to create a more nuanced balance by generating new, synthetic samples, yet these might not perfectly capture the underlying complexities of the data. QI-SMOTE's approach, by generating samples that potentially represent the data complexity more accurately, can offer a more refined solution in maintaining the integrity and diversity of the dataset, which is particularly important in complex domains like healthcare. Expanding our analysis to incorporate a third feature dimension with three-dimensional scatter plots, as shown in figure 3, provided an even more nuanced view of the data distribution. In this more complex visualization, it was evident that the synthetic samples introduced by QI-SMOTE were uniformly dispersed throughout the dataset. This uniform interspersion is indicative of QI-SMOTE's capability to create a balanced dataset, one that doesn't exhibit a bias towards specific regions or classes. Such a balanced distribution is crucial for training ML models that are robust and unbiased, especially in scenarios where class imbalance can significantly skew predictions.

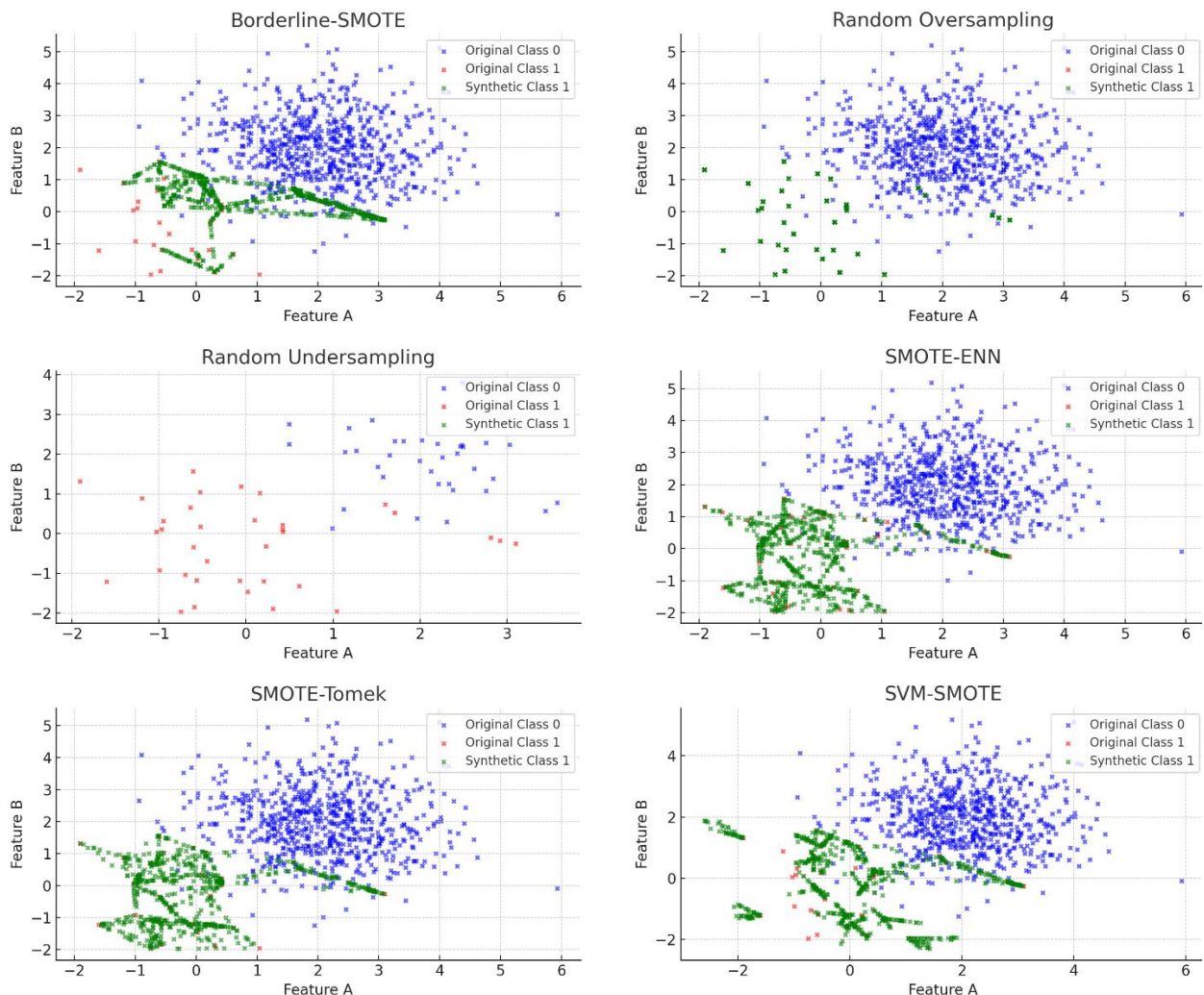

**Figure 2.** *(continued)*.

## 5.2 Quantitative Results

Analysing the F-measure data across MIMIC-III and MIMIC-IV datasets as shown in the Figure 4 and 5 respectively, reveals the effectiveness of different oversampling techniques in handling class imbalance. QI-SMOTE stands out consistently, showing high F-measure scores across multiple classifiers such as DT, GB, KNN, and RF. This indicates its robustness in creating balanced datasets that enhance algorithm performance. In contrast, the original dataset without oversampling typically displays lower F-measure scores. This discrepancy becomes more pronounced in datasets with greater imbalance, like MIMIC-III (10) and MIMIC-III (20), where scores for methods like ADA, LR, and SVM particularly decline. Other techniques like SMOTE, ADASYN, B-SMOTE, and ROS demonstrate varied performance. For example, SMOTE improves in classifiers like ADA and LR from MIMIC-III to MIMIC-III (10), although not as consistently as QI-SMOTE. As class imbalance intensifies in datasets such as MIMIC-III (10) and MIMIC-III (20), a general decrease in F-measure is observed for most techniques and classifiers, highlighting the complexities of handling imbalance. QI-SMOTE, however, maintains relatively higher scores, underscoring its effectiveness. In the MIMIC-IV dataset, QI-SMOTE again delivers high F-measure scores, demonstrating its capability to manage varying levels of class imbalance and dataset complexity effectively. The analysis emphasizes QI-SMOTE's ability to maintain a balance between precision and recall, especially in complex scenarios like MIMIC-IV (20).

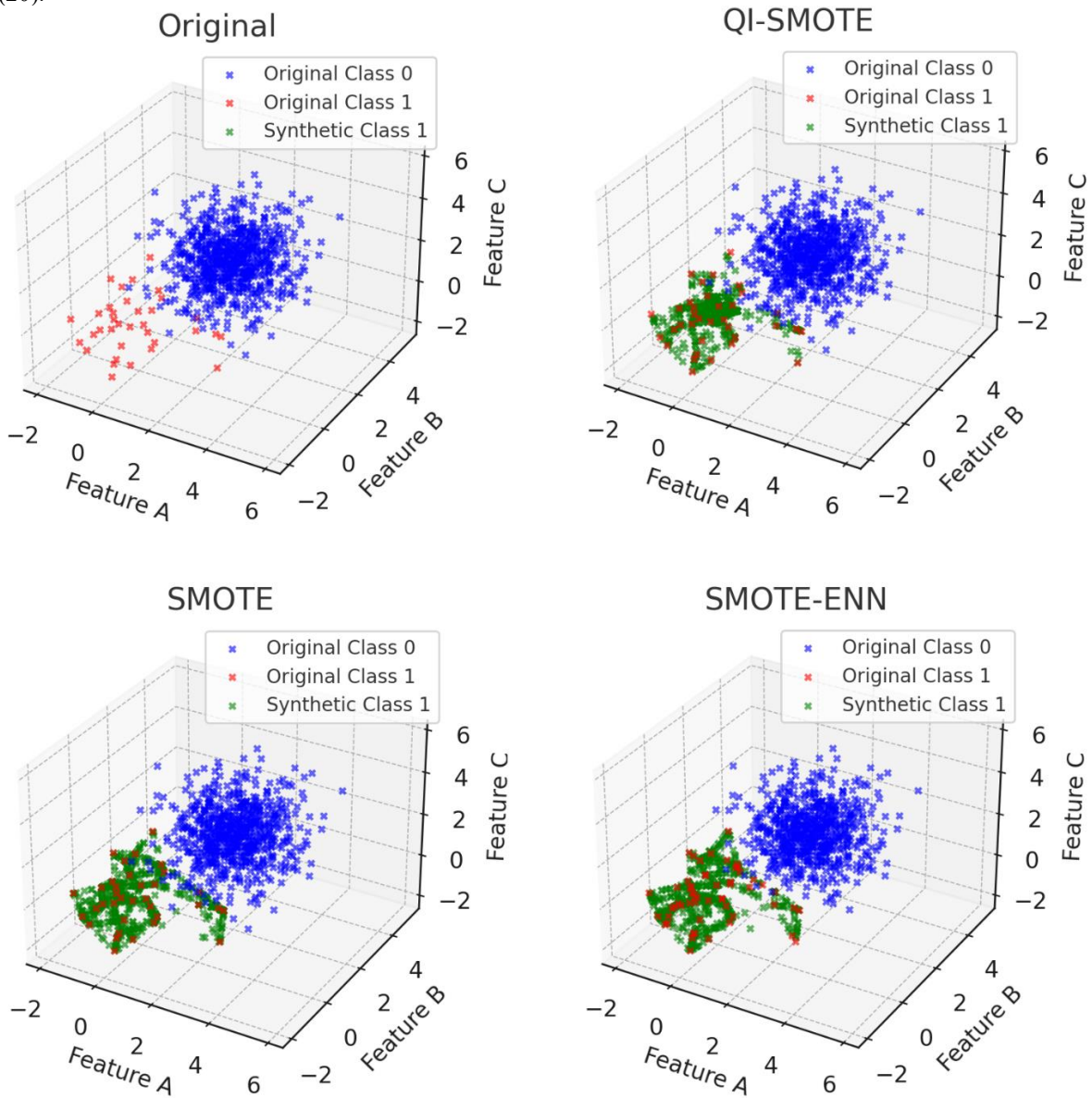

**Figure 3.** Three-dimensional scatter plots illustrating the effect of QI-SMOTE and other oversampling methods in a higher-dimensional space. The QI-SMOTE-generated samples exhibit more uniform dispersion and coverage of the minority class.

Table 1. shows percentage improvement of F1-score over the original dataset for each technique across different MIMIC variants as per equation (1) below:

$$\text{Improvement}(\%) = \frac{F1_{technique} - F1_{original}}{F1_{original}} \times 100 \quad (1)$$

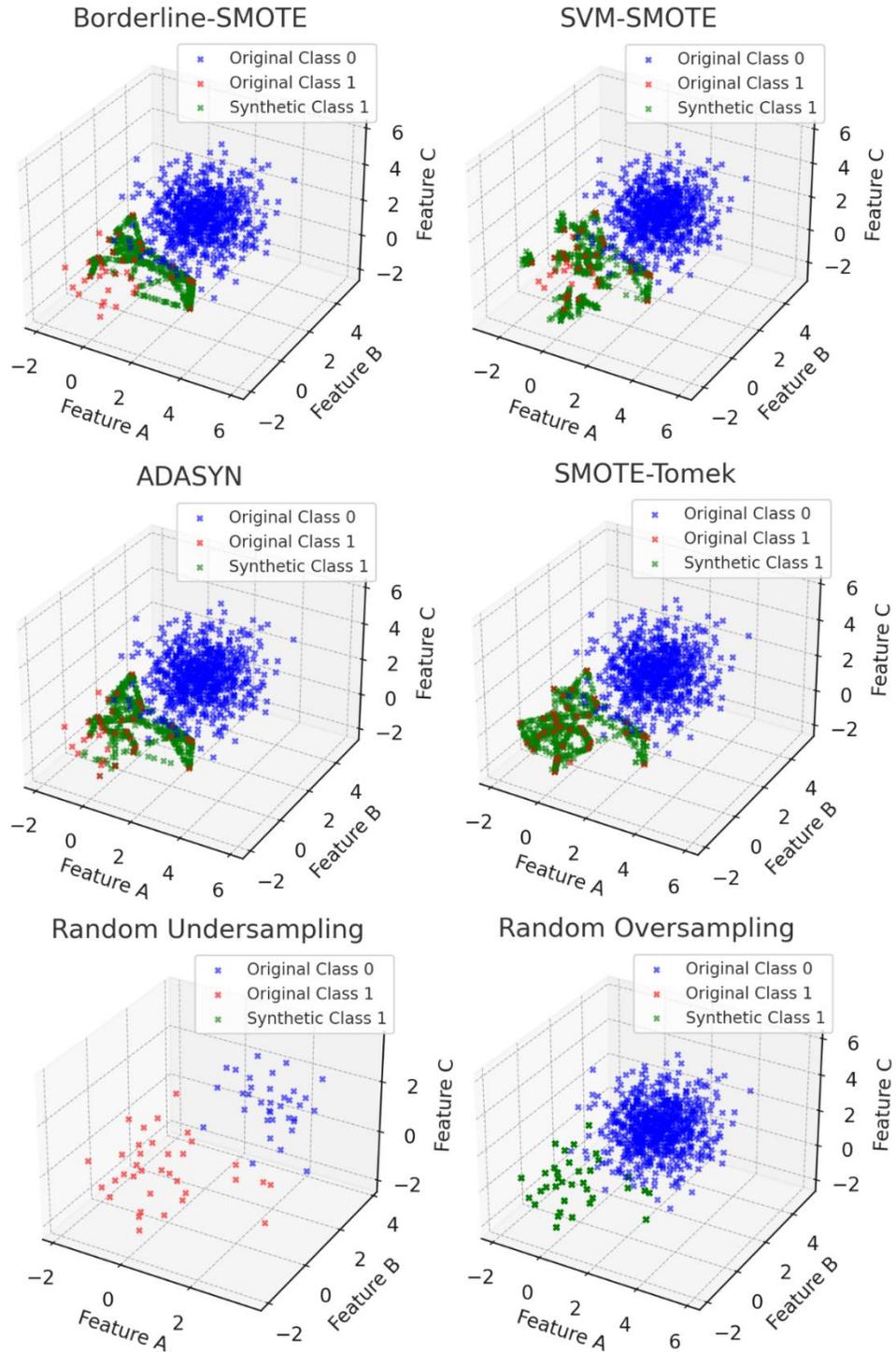

**Figure 3.** *(continued)*.

For the MIMIC-III dataset, QI-SMOTE shows a significant improvement of 12.11% over the original dataset, surpassing other methods such as SMOTE, ADASYN, and B-SMOTE, which report improvements of 2.81%, 5.02%, and 3.84%. This trend is consistent across variations of the dataset, with QI-SMOTE maintaining a leading edge in the more challenging scenarios (MIMIC-III (20)) with an 11.22% improvement, underscoring its robustness in handling varying degrees of class imbalance. The performance leap is even more pronounced within the MIMIC-IV dataset, where QI-SMOTE achieves a remarkable 24.30% improvement over the original dataset, significantly outpacing the nearest competitor, SMOTE-ENN, which achieves a 22.38% improvement. The disparity in performance becomes stark in the extended scenarios (MIMIC-IV (10) and MIMIC-IV (20)), where QI-SMOTE demonstrates exponential improvements of 89.52% and 165.28%, respectively. This is indicative of QI-SMOTE's exceptional ability to generate synthetic instances that effectively mimic the complex, underlying data structure of medical datasets, thereby enhancing model performance substantially. Comparatively, traditional methods like SMOTE, ADASYN, and B-SMOTE show moderate improvements, with their effectiveness tapering off in more complex scenarios. We observe that in the MIMIC-III dataset QI-SMOTE exhibits a strong performance with an accuracy of around 0.755. This performance is consistent even as the imbalance increases in MIMIC-III (10) and MIMIC-III (20) as shown in Table 2. The Original dataset, without oversampling, shows significantly lower performance metrics across all versions, indicating the necessity of oversampling techniques in managing class imbalance.

**Table 1.** Percentage improvement of F1 over the original dataset.

| Technique | MIMIC-III | MIMIC-III (10) | MIMIC-III (20) | MIMIC-IV | MIMIC-IV (10) | MIMIC-IV (20) |
|---|---|---|---|---|---|---|
| QI-SMOTE | 12.11% | 5.86% | 11.22% | 24.30% | 89.52% | 165.28% |
| SMOTE | 2.81% | -10.36% | 8.01% | 9.97% | 75.00% | 144.91% |
| ADASYN | 5.02% | -9.46% | 1.28% | 13.29% | 75.54% | 149.06% |
| B-SMOTE | 3.84% | 1.65% | 8.17% | 15.38% | 73.12% | 137.36% |
| ROS | 5.17% | 5.41% | 8.65% | 9.79% | 86.56% | 82.26% |
| RUS | 1.62% | 4.95% | -35.58% | 17.13% | 85.75% | 105.28% |
| SMOTE-ENN | -1.33% | 1.35% | 5.61% | 22.38% | 84.95% | 155.85% |
| SMOTE-TOMEK | 0.30% | 1.50% | 7.85% | 9.62% | 75.00% | 144.91% |
| SVM-SMOTE | 0.15% | 2.40% | 7.85% | 11.19% | 58.60% | 63.77% |

Moving to the MIMIC-IV dataset, designed to represent a more advanced and potentially more complex data structure, we see a similar trend. QI-SMOTE continues to demonstrate robust performance, indicating its effectiveness in dealing with complex data scenarios as shown in Table 3. The Original dataset again shows lower performance metrics, reinforcing the challenges posed by class imbalance. When comparing techniques like SMOTE and SVM-SMOTE across these datasets, there's a noticeable variability in their performance. In summary, the analysis of MIMIC-III and MIMIC-IV datasets suggests that while all techniques are impacted by increasing class imbalance and data complexity, certain methods like QI-SMOTE effective in maintaining balanced performance. The improvement in metrics, especially the F1-Score and G-Mean, highlight the potential of QI-SMOTE in enhancing the classification performance of ML models in the medical domain.

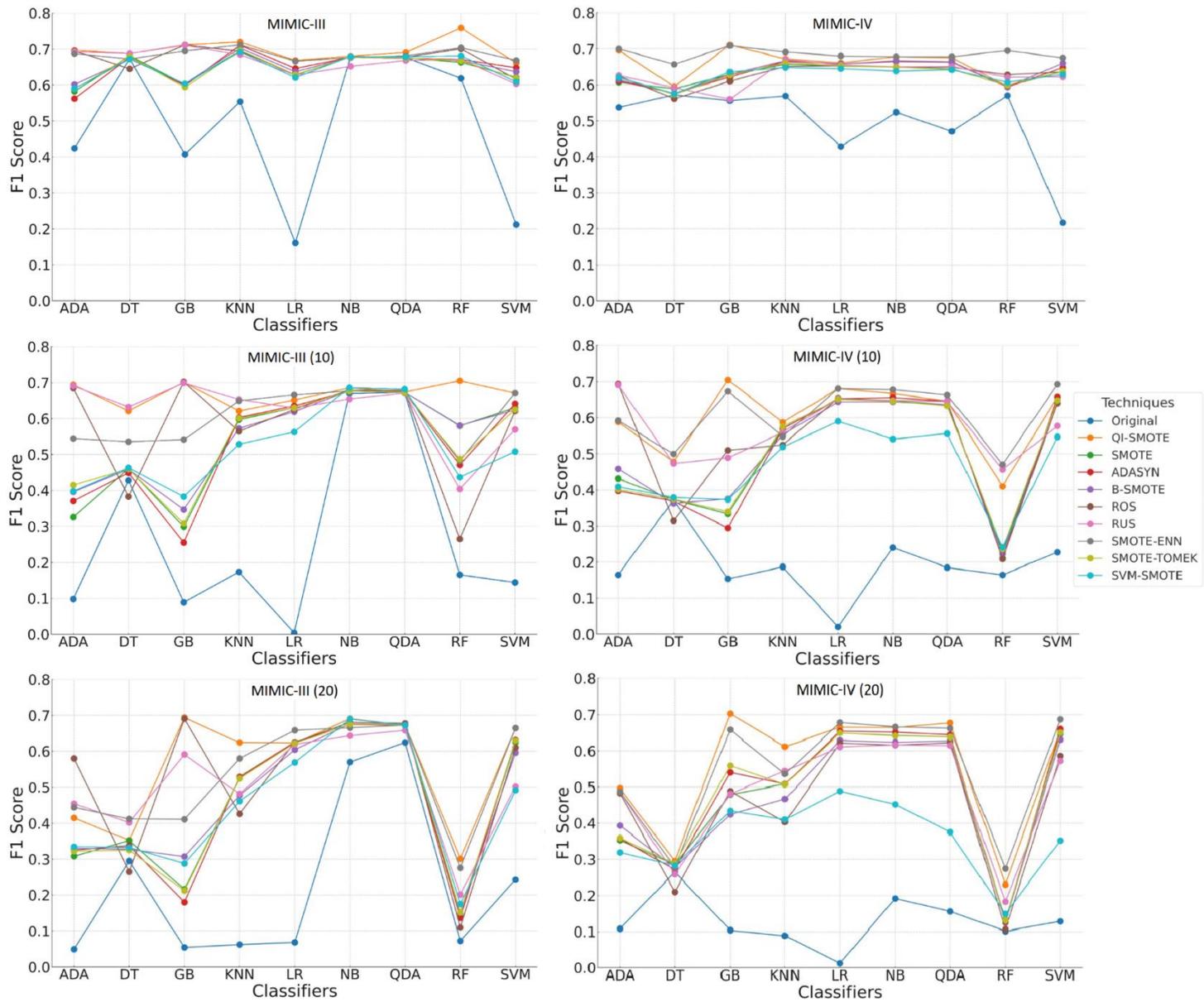

**Figure 4.** Comparison of F1 Score of each oversampling algorithm on MIMIC-III and MIMIC-IV Datasets and their imbalanced variants

**Table 2.** Performance comparison of different oversampling techniques on the MIMIC-III dataset and its variants.

| Dataset | Technique | Accuracy | G-Mean | AUC-Score | Precision | Recall | F1 |
|---|---|---|---|---|---|---|---|
| MIMIC-III | Original | 0.580 | 0.494 | 0.684 | 0.550 | 0.886 | 0.677 |
| | QI-SMOTE | 0.755 | 0.755 | 0.838 | 0.748 | 0.770 | 0.759 |
| | SMOTE | 0.677 | 0.741 | 0.690 | 0.737 | 0.659 | 0.696 |
| | RUS | 0.748 | 0.598 | 0.836 | 0.672 | 0.705 | 0.688 |
| | ROS | 0.713 | 0.724 | 0.740 | 0.655 | 0.781 | 0.712 |
| | ADASYN | 0.630 | 0.534 | 0.768 | 0.721 | 0.701 | 0.711 |
| | B-SMOTE | 0.729 | 0.650 | 0.749 | 0.722 | 0.685 | 0.703 |
| | SVM-SMOTE | 0.595 | 0.613 | 0.828 | 0.696 | 0.661 | 0.678 |
| | SMOTE-ENN | 0.619 | 0.738 | 0.816 | 0.704 | 0.636 | 0.668 |
| | TOMEK-SMOTE | 0.641 | 0.734 | 0.807 | 0.683 | 0.675 | 0.679 |
| MIMIC-III (10) | Original | 0.550 | 0.485 | 0.677 | 0.530 | 0.893 | 0.666 |
| | QI-SMOTE | 0.738 | 0.708 | 0.831 | 0.732 | 0.680 | 0.705 |
| | SMOTE | 0.599 | 0.638 | 0.695 | 0.618 | 0.577 | 0.597 |
| | RUS | 0.627 | 0.663 | 0.721 | 0.636 | 0.776 | 0.699 |
| | ROS | 0.613 | 0.614 | 0.774 | 0.661 | 0.748 | 0.702 |
| | ADASYN | 0.741 | 0.704 | 0.836 | 0.662 | 0.554 | 0.603 |
| | B-SMOTE | 0.609 | 0.625 | 0.781 | 0.704 | 0.652 | 0.677 |
| | SVM-SMOTE | 0.717 | 0.624 | 0.712 | 0.651 | 0.716 | 0.682 |
| | SMOTE-ENN | 0.646 | 0.577 | 0.824 | 0.677 | 0.673 | 0.675 |
| | TOMEK-SMOTE | 0.608 | 0.502 | 0.753 | 0.745 | 0.619 | 0.676 |
| MIMIC-III (20) | Original | 0.545 | 0.480 | 0.671 | 0.528 | 0.763 | 0.624 |
| | QI-SMOTE | 0.701 | 0.743 | 0.833 | 0.736 | 0.657 | 0.694 |
| | SMOTE | 0.600 | 0.733 | 0.823 | 0.585 | 0.795 | 0.674 |
| | RUS | 0.607 | 0.574 | 0.817 | 0.684 | 0.285 | 0.402 |
| | ROS | 0.673 | 0.745 | 0.706 | 0.662 | 0.695 | 0.678 |
| | ADASYN | 0.636 | 0.744 | 0.804 | 0.610 | 0.656 | 0.632 |
| | B-SMOTE | 0.697 | 0.729 | 0.697 | 0.684 | 0.666 | 0.675 |
| | SVM-SMOTE | 0.644 | 0.498 | 0.738 | 0.660 | 0.687 | 0.673 |
| | SMOTE-ENN | 0.589 | 0.612 | 0.729 | 0.680 | 0.639 | 0.659 |
| | TOMEK-SMOTE | 0.607 | 0.578 | 0.816 | 0.701 | 0.647 | 0.673 |

**Table 3.** Performance comparison of different oversampling techniques on the MIMIC-IV dataset and its variants

| Dataset | Technique | Accuracy | G-Mean | AUC-Score | Precision | Recall | F1 |
|---|---|---|---|---|---|---|---|
| MIMIC-IV | Original | 0.600 | 0.514 | 0.704 | 0.570 | 0.574 | 0.572 |
| | QI-SMOTE | 0.755 | 0.745 | 0.858 | 0.758 | 0.669 | 0.711 |
| | SMOTE | 0.688 | 0.590 | 0.737 | 0.606 | 0.654 | 0.629 |
| | RUS | 0.620 | 0.725 | 0.740 | 0.679 | 0.661 | 0.670 |
| | ROS | 0.680 | 0.539 | 0.708 | 0.648 | 0.609 | 0.628 |
| | ADASYN | 0.634 | 0.668 | 0.854 | 0.572 | 0.747 | 0.648 |
| | B-SMOTE | 0.741 | 0.630 | 0.847 | 0.638 | 0.684 | 0.660 |
| | SVM-SMOTE | 0.703 | 0.683 | 0.721 | 0.593 | 0.686 | 0.636 |
| | SMOTE-ENN | 0.721 | 0.685 | 0.736 | 0.599 | 0.842 | 0.700 |
| | TOMEK-SMOTE | 0.658 | 0.721 | 0.817 | 0.750 | 0.539 | 0.627 |
| MIMIC-IV (10) | Original | 0.592 | 0.505 | 0.681 | 0.565 | 0.277 | 0.372 |
| | QI-SMOTE | 0.771 | 0.768 | 0.859 | 0.767 | 0.652 | 0.705 |
| | SMOTE | 0.760 | 0.717 | 0.720 | 0.715 | 0.598 | 0.651 |
| | RUS | 0.708 | 0.754 | 0.801 | 0.628 | 0.768 | 0.691 |
| | ROS | 0.601 | 0.574 | 0.782 | 0.687 | 0.701 | 0.694 |
| | ADASYN | 0.709 | 0.647 | 0.715 | 0.706 | 0.607 | 0.653 |

|  | B-SMOTE | 0.666 | 0.725 | 0.782 | 0.651 | 0.637 | 0.644 |
|  | SVM-SMOTE | 0.707 | 0.746 | 0.796 | 0.609 | 0.572 | 0.590 |
|  | SMOTE-ENN | 0.617 | 0.567 | 0.782 | 0.589 | 0.829 | 0.688 |
|  | TOMEK-SMOTE | 0.606 | 0.747 | 0.857 | 0.663 | 0.639 | 0.651 |
| MIMIC-IV (20) | Original | 0.589 | 0.500 | 0.678 | 0.561 | 0.173 | 0.265 |
|  | QI-SMOTE | 0.771 | 0.765 | 0.849 | 0.767 | 0.649 | 0.703 |
|  | SMOTE | 0.646 | 0.515 | 0.796 | 0.754 | 0.570 | 0.649 |
|  | RUS | 0.641 | 0.743 | 0.815 | 0.753 | 0.426 | 0.544 |
|  | ROS | 0.673 | 0.717 | 0.819 | 0.757 | 0.355 | 0.483 |
|  | ADASYN | 0.727 | 0.541 | 0.723 | 0.656 | 0.664 | 0.660 |
|  | B-SMOTE | 0.602 | 0.541 | 0.844 | 0.715 | 0.561 | 0.629 |
|  | SVM-SMOTE | 0.627 | 0.631 | 0.804 | 0.763 | 0.303 | 0.434 |
|  | SMOTE-ENN | 0.719 | 0.553 | 0.754 | 0.662 | 0.695 | 0.678 |
|  | TOMEK-SMOTE | 0.601 | 0.754 | 0.798 | 0.680 | 0.621 | 0.649 |

## 5.3 Statistical Significance Analysis

We conducted statistical significance testing on F1-score to verify the robustness of QI-SMOTE's performance improvements over classical oversampling techniques. We chose to employ the Wilcoxon signed-rank test, a non-parametric alternative to the paired t-test because this method does not assume normality in the distribution of score differences, making it well-suited for small sample sizes and real-world clinical data with unknown or non-Gaussian characteristics. To assess the statistical significance of the performance differences between QI-SMOTE and SMOTE-ENN, a two-sided Wilcoxon signed-rank test (paired) was applied to their F1 scores obtained over ten cross-validation folds on MIMIC-IV (10) dataset. The F1 performance differences were calculated as shown in Table 4 and then ranks were calculated as shown in Table 5. The effect size $r = |z|/\sqrt{N}$ along with $W / Z / p$ were reported in the Table 6. Positive Δ% indicates improvement over the baseline.

**Table 4.** Paired F1 scores for QI-SMOTE and SMOTE-ENN across 10 cross-validation folds, along with the computed difference (Difference=QI-SMOTE−SMOTE-ENN). Difference forms the basis for the Wilcoxon Signed-Rank Test

| Fold | QI-SMOTE | SMOTE-ENN | Difference ($d_i$) |
|---|---|---|---|
| 1 | 0.782 | 0.763 | 0.019 |
| 2 | 0.776 | 0.763 | 0.013 |
| 3 | 0.783 | 0.770 | 0.013 |
| 4 | 0.792 | 0.749 | 0.043 |
| 5 | 0.775 | 0.751 | 0.024 |
| 6 | 0.775 | 0.762 | 0.013 |
| 7 | 0.793 | 0.758 | 0.035 |
| 8 | 0.785 | 0.771 | 0.014 |
| 9 | 0.772 | 0.759 | 0.013 |
| 10 | 0.782 | 0.754 | 0.028 |

**Table 5.** Absolute paired differences and corresponding rank assignments used in the Wilcoxon Signed-Rank Test. Absolute differences were sorted in ascending order, and average ranks were assigned to tied values. These ranks reflect the relative magnitude of each performance difference.

| Fold | Difference ($d_i$) | Rank |
|---|---|---|
| 2 | 0.013 | 2.500 |
| 3 | 0.013 | 2.500 |
| 6 | 0.013 | 2.500 |
| 9 | 0.013 | 2.500 |
| 8 | 0.014 | 5.000 |

| 1 | 0.019 | 6.000 |
| 5 | 0.024 | 7.000 |
| 10 | 0.028 | 8.000 |
| 7 | 0.035 | 9.000 |
| 4 | 0.043 | 10.000 |

Since all differences are positive, all signed ranks are positive.

Signed Rank$_i$ = sign($d_i$) × Rank$_i$

$W^+ = \sum$ (Positive Signed Ranks) = 2.5 + 2.5 + 2.5 + 2.5 + 5 + 6 + 7 + 8 + 9 + 10 = 55.0

$W^- = 0$   (since no negative differences)

$W = \min(W^+, W^-) = 0$

Table 6. Wilcoxon signed-rank test summary

| Comparison | N (pairs) | T+ (Sum of positive ranks) | T− (Sum of negative ranks) | W (Min of T+, T−) | z | p (two-sided, exact) | Effect size $r = \|z\|/\sqrt{N}$ |
|---|---|---|---|---|---|---|---|
| QI-SMOTE vs SMOTE-ENN | 10 | 55 | 0 | 0 | 2.891 | 0.001953 | 0.914 |

The p-value was computed using the exact distribution of the Wilcoxon statistic for $n = 10$, based on enumeration of all 2^10 = 1024 signed rank permutations. Only one permutation yields $W = 0$, resulting in a p-value of 2/1024 for two-sided Wilcoxon signed-rank test, which is 0.001953, indicating a highly significant result. This result indicates that QI-SMOTE consistently outperformed SMOTE-ENN across all folds, with no instances of performance degradation. The value of W was calculated to be 0 which denotes the strongest possible deviation from the null hypothesis, suggesting that the observed improvements are highly unlikely to have occurred by chance. Thus, the results provide strong evidence that QI-SMOTE offers a statistically significant enhancement in classification performance over the SMOTE-ENN technique.

### 5.4 Qualitative Analysis

During exploratory analysis on the MIMIC-IV dataset, we found that traditional SMOTE occasionally produced synthetic minority samples with implausible physiological combinations, such as systolic blood pressure approaching 160 mmHg while heart rate remained below 40 bpm as shown in Figure 5. While rare, such outliers can skew classifier boundaries and contribute to false positives. In contrast, QI-SMOTE's entanglement mechanism ensured that features like blood pressure and heart rate were co-modulated, yielding synthetic samples that stayed within physiologically realistic bounds. This is reflected in improved classifier calibration and F1-score stability, especially under higher imbalance settings (MIMIC-IV (10) and (20)).

The experimental evaluation of QI-SMOTE on the MIMIC-III and MIMIC-IV datasets provided valuable insights into the algorithm's strengths and weaknesses.

**Strengths:** On MIMIC-III and MIMIC-IV, QI-SMOTE consistently exceeded SMOTE-family baselines (e.g., SMOTE-ENN, ADASYN, Borderline-SMOTE) on F1, G-Mean, and AUC, and stayed robust across different classifiers and imbalance levels. By preserving high-dimensional structure (via its quantum-inspired embedding) it generates synthetics closer to the minority support, improving plausibility and reducing overfitting.

**Weaknesses:** The method is more compute- and memory-intensive than classical oversampling methods, and its quantum-inspired components add hyperparameters that require careful tuning. As with any oversampling approach, synthetic quality is limited by the source data—noisy or biased minority samples can propagate into the generated points.

In conclusion, while QI-SMOTE presents an effective approach to address the class imbalance problem, especially in the context of medical data, it is essential to consider its computational demands and the quality of the original data when deploying it in real-world scenarios.

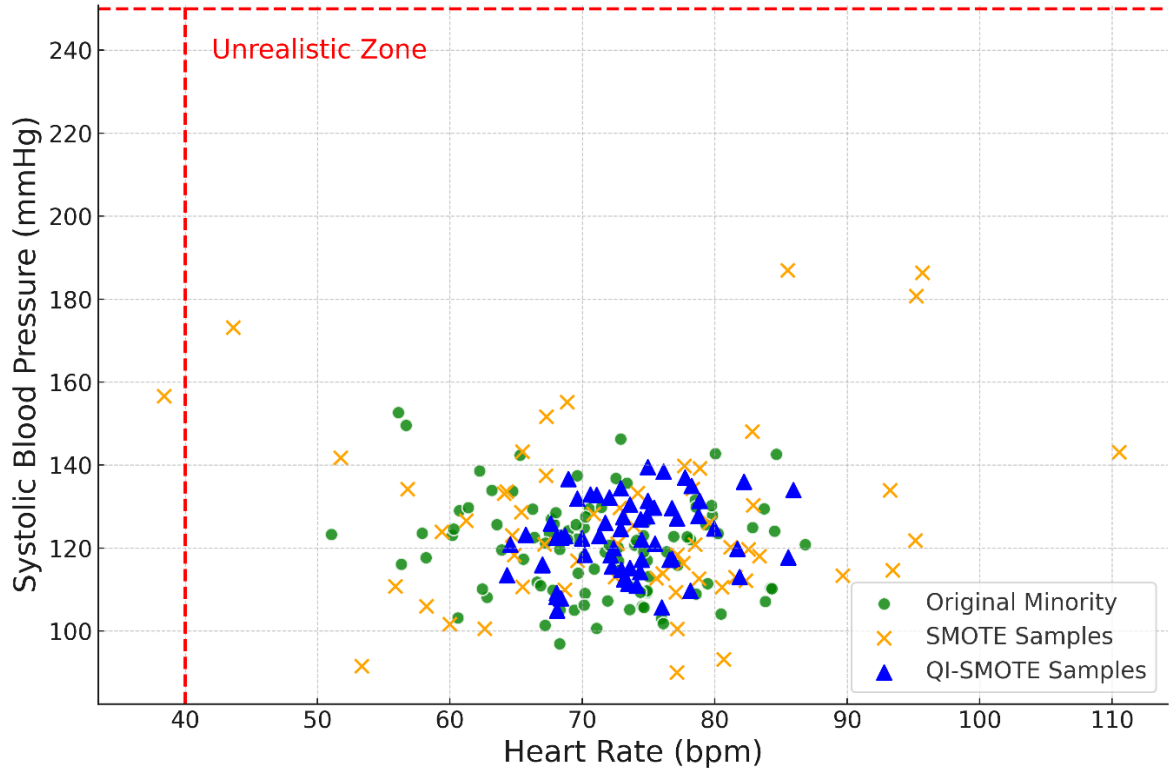

**Figure 5.** Synthetic Sample Distribution — SMOTE vs. QI-SMOTE on MIMIC-IV (10)
This figure compares synthetic sample distributions generated by classical SMOTE and QI-SMOTE in a 2D PCA projection space. Green points represent original minority class samples. SMOTE-generated samples (orange) tend to extrapolate into less realistic zones of the feature space, while QI-SMOTE-generated samples (blue) remain more densely clustered around the minority manifold. The red dotted line marks a region where classical oversampling risks producing unrealistic combinations, whereas QI-SMOTE preserves neighbourhood geometry through quantum-evolved features.

### 5.5 Runtime and Memory Usage Benchmark

To quantify the computational cost, we benchmarked both SMOTE and QI-SMOTE on a dataset of 700 samples with 3 features each, where 30 samples belonged to the minority class. Using quantum hyperparameters as stated in Appendix A, we generated 100 additional synthetic minority samples using both methods. On a standard Intel i7 CPU with 32GB RAM, traditional SMOTE completed this task in under 0.1 seconds and used less than 1MB of RAM. In contrast, QI-SMOTE simulated on the same hardware using Qiskit, required approximately 50-100 seconds and up to 200MB of RAM for the same operation. This significant increase in computational resources is due to the overhead from simulating quantum circuits and running the iterative VQE optimization, even in this low-dimensional, tabular setting with just three features.

The higher runtime and memory use of QI-SMOTE today is mainly because it relies on simulating quantum circuits on regular computers, which is much slower than running classical algorithms. However, this is expected to change as quantum computers become available. Once actual quantum hardware can be used, quantum-inspired algorithms like QI-SMOTE will run much faster and more efficiently. This means that, while the extra cost is a trade-off today, it is likely to disappear in the near future. By developing and testing these advanced methods now, we are preparing for a time when quantum computing will make such powerful algorithms practical for real-world, time-sensitive tasks.

### 5.6 Implications

**Clinical impact:** Better handling of imbalance can raise diagnostic/prognostic accuracy and support more informed decisions. Stronger models also enable personalized medicine by tailoring interventions to individual profiles.

**Scalability & transfer:** Demonstrated on high-dimensional MIMIC-III/IV, QI-SMOTE shows promise for big-data settings and can be adapted to other imbalanced domains (e.g., finance, cybersecurity, environmental monitoring).

**Methodological advance:** Bridges quantum principles with classical ML, encouraging further quantum-inspired algorithms and hybrid approaches.

**Responsible deployment:** Requires safeguards against bias amplification from synthetic data and strict protection of patient privacy and broader ethical considerations.

In essence, the development and validation of QI-SMOTE not only address a technical challenge in ML but also open doors to transformative changes in how we harness data for the betterment of society.

# 6 LIMITATIONS

Despite the promising results of QI-SMOTE in enhancing classification performance for imbalanced clinical datasets, there are some limitations worth noting. First, the use of quantum-inspired simulations, such as entanglement and VQE optimization-introduces computational overhead that may not be scalable to extremely large datasets without further optimization. Second, while we observe consistent gains across multiple metrics, the method involves several hyperparameters (e.g., entanglement depth, number of VQE iterations) that may require tuning in other domains or datasets. Third, as the current implementation simulates quantum operations classically, the hardware-specific behaviours, noise effects, or scalability under real quantum environments remain unexplored. Lastly, the method's generalizability beyond structured clinical tabular data has not yet been evaluated. Extensive research is needed to explore hybrid quantum-classical implementations, hyperparameter robustness across domains, and extensions to multimodal or time-series datasets.

# 7 CONCLUSION AND FUTURE WORK

In this study, we introduced QI-SMOTE, a quantum-inspired approach designed to enhance ML techniques by effectively addressing the class imbalance problem in multidimensional medical data. Our experimental results show that QI-SMOTE outperforms conventional resampling methods in improving classifier performance, particularly in handling complex medical datasets. Its adaptability to various data distributions, along with its ability to reduce overfitting and enhance model generalization, underscores its significance in ML-based predictive analytics. Beyond empirical improvements, QI-SMOTE contributes an innovative formulation to the resampling literature by integrating quantum mechanics–inspired concepts, namely superposition, entanglement, and Hamiltonian-based evolution, into the synthetic data generation process. Unlike kernel-based or autoencoder-based oversampling methods, which seek nonlinear manifold representations through abstract mathematical mappings, QI-SMOTE employs physically interpretable operations that simulate quantum behavior. To the best of our knowledge, the use of a VQE loop to optimize feature states prior to classical SMOTE interpolation is a unique contribution. While this approach is implemented entirely in classical simulation and does not target quantum speedup, it represents a creative and effective strategy for structure-aware augmentation in complex domains like medical diagnostics.

Currently, QI-SMOTE is implemented as a classical simulation of quantum-inspired operations, which can be computationally demanding. As practical, large-scale quantum computers become available, there is the potential for substantial speedup and efficiency gains in executing the quantum components of our approach. Beyond its impact on medical AI, QI-SMOTE has the potential to benefit a wide range of ML applications in domains such as finance, cybersecurity, and e-commerce, where class imbalance remains a persistent challenge. By effectively bridging the gap between quantum computing and ML, QI-SMOTE not only advances ML methodologies but also paves the way for novel approaches in data augmentation and model optimization. We are optimistic that the fusion of quantum principles with ML holds immense potential for transformative innovations.

**APPENDIX A:** Quantum Gate Definitions and Operations

This appendix provides a reference overview of the fundamental quantum gates and operations used in the quantum-inspired architecture of QI-SMOTE. These gates simulate how classical features are transformed into entangled quantum-like states prior to synthetic data generation.

### A.1 Hadamard Gate (H)

The Hadamard gate is a single-qubit gate used to place a qubit into a superposition state. It transforms a basis state $|0\rangle$ or $|1\rangle$ into an equal-weighted combination of $|0\rangle$ and $|1\rangle$, enabling probabilistic representation of feature values.

$$H = \frac{1}{\sqrt{2}} \begin{bmatrix} 1 & 1 \\ 1 & -1 \end{bmatrix} \quad (A.1.1)$$

$$H|0\rangle = \frac{1}{\sqrt{2}}(|0\rangle + |1\rangle), \quad H|1\rangle = \frac{1}{\sqrt{2}}(|0\rangle - |1\rangle) \quad (A.1.2)$$

In QI-SMOTE, Hadamard gates are used to initialize each qubit in a state of superposition, allowing parallel encoding of feature dynamics.

### A.2 RY Rotation Gate

The RY gate performs a rotation around the Y-axis of the Bloch sphere. It is parameterized by an angle $\theta$, which is derived from a normalized feature value $f_i$. This mapping transforms classical input into a continuous quantum state representation.

$$RY(\theta) = \begin{bmatrix} \cos(\theta/2) & -\sin(\theta/2) \\ \sin(\theta/2) & \cos(\theta/2) \end{bmatrix} \quad (A.2.1)$$

Before applying the RY rotation gate to each qubit, the raw feature values $f_i$ are min-max normalized into the range $[0, \pi]$ to ensure valid rotation angles:

$$\theta_i = \pi \cdot \frac{f_i - \min(f)}{\max(f) - \min(f)} \quad (A.2.2)$$

The RY gate is then applied to the qubit initialized by the Hadamard gate, encoding the feature value into qubit amplitude.

### A.3 Controlled-NOT Gate (CNOT)

The CNOT gate entangles two qubits by flipping the target qubit if the control qubit is in state $|1\rangle$ This operation is essential for modeling pairwise dependencies between features.

$$\text{CNOT} = \begin{bmatrix} 1 & 0 & 0 & 0 \\ 0 & 1 & 0 & 0 \\ 0 & 0 & 0 & 1 \\ 0 & 0 & 1 & 0 \end{bmatrix} \quad (A.3.1)$$

The CNOT gate ensures conditional logic: if Feature A influences Feature B in the training data, their quantum analogues become entangled via this gate.

### A.4 Controlled-Z Gate (CZ)

The CZ gate introduces a conditional phase shift, flipping the phase of the target qubit when the control qubit is $|1\rangle$. This gate enables phase entanglement between features, preserving certain covariance relationships in high-dimensional distributions.

$$\text{CZ} = \begin{bmatrix} 1 & 0 & 0 & 0 \\ 0 & 1 & 0 & 0 \\ 0 & 0 & 1 & 0 \\ 0 & 0 & 0 & -1 \end{bmatrix} \quad (A.4.1)$$

The CZ gate allows for modeling phase-based interactions not captured by amplitude changes alone.

### A.5 Toffoli Gate (CCNOT)

The Toffoli gate is a three-qubit gate that flips the target qubit only if both control qubits are in state $|1\rangle$. This gate introduces non-linear interactions and high-order correlations across multiple features.

$$\text{Toffoli}(a,b,c) = \begin{cases} c \to \neg c & \text{if } a = b = 1 \\ c \to c & \text{otherwise} \end{cases} \quad (A.5.1)$$

This operation captures complex structures such as logical conjunctions between multiple dependent medical variables (e.g., blood pressure, heart rate, and oxygen saturation).

### A.6 Hamiltonian Encoding and Variational Quantum Eigensolver (VQE) Optimization

Once the qubits are initialized, rotated, and entangled, a Hamiltonian is defined to represent the energy landscape of the synthetic feature state:

$$H = \sum_{i,j} w_{ij} Z_i Z_j + \sum_i b_i Z_i \quad (A.6.1)$$

Here, $Z_i$ are Pauli-Z operators, and $w_{ij}, b_i$ are learned weights reflecting the coupling between features. This Hamiltonian resembles an Ising-type energy landscape, which is often used in quantum machine learning for capturing both individual qubit contributions and pairwise correlations.

A VQE is then used to find the optimal set of parameters $\theta^*$ that minimizes the expected energy of the system:

$$\theta^* = \arg\min_\theta \langle \psi(\theta) | H | \psi(\theta) \rangle \quad (A.6.2)$$

For the VQE optimization step, the following hyperparameters were used across all datasets:

**Optimizer:** COBYLA

**Maximum Iterations:** 100 per sample

**Convergence Tolerance:** $10^{-6}$

**Ansatz Structure:** 2 entangling layers using RY rotations and CZ gates

This ensures that the synthetic data resides in a low-energy, highly structured region of feature space, avoiding noisy or unrepresentative samples.